%% file: main.tex
\documentclass[10pt,twocolumn,letterpaper]{article}

\usepackage{cvpr}              
\usepackage{csquotes}
\usepackage{algorithm}
\usepackage{algpseudocode}
\usepackage{multirow}

\usepackage{textcomp} 
\usepackage{marvosym}
\usepackage{colortbl}
\usepackage{xcolor}
\usepackage{diagbox}
\usepackage{wasysym}
\usepackage{marvosym}
\usepackage{pifont}
\usepackage{setspace}

\usepackage{enumitem}

\usepackage{makecell}

\input{preamble}

\definecolor{cvprblue}{rgb}{0.21,0.49,0.74}
\definecolor{gcol}{RGB}{70,0,120}
\definecolor{ccol}{RGB}{90,0,150}
\definecolor{scol}{RGB}{110,0,180}
\definecolor{ecol}{RGB}{130,20,200}
\definecolor{dcol}{RGB}{150,40,215}
\definecolor{icol}{RGB}{170,70,225}
\definecolor{tcol}{RGB}{190,100,235}

\usepackage[pagebackref,breaklinks,colorlinks,allcolors=cvprblue]{hyperref}
\usepackage[most,theorems,skins,listings]{tcolorbox}

\def\paperID{5447} 
\def\confName{CVPR}
\def\confYear{2026}

\title{
\textcolor{gcol}{L}%
\textcolor{ccol}{i}%
\textcolor{scol}{v}%
\textcolor{ecol}{e}%
\textcolor{dcol}{E}%
\textcolor{icol}{d}%
\textcolor{tcol}{i}%
\textcolor{tcol}{t}:
Towards Real-Time Diffusion-Based Streaming Video Editing
}
\author{%
  Xinyu Wang\textsuperscript{1},
  Chongbo Zhao\textsuperscript{1},
  Fangneng Zhan\textsuperscript{2},
  Yue Ma\textsuperscript{2}\textsuperscript{\Letter}
  \\[1mm]
  \textsuperscript{1} THU \quad 
  \textsuperscript{2} HKUST \quad
   \\
  \\
  \textbf{Project: \href{https://live-edit.github.io}{\texttt{\textcolor{cyan}{https://live-edit.github.io}}}}
}

\begin{document}

\twocolumn[{
\begin{center}
\maketitle
\vspace{-2em}
    \captionsetup{type=figure}
    \includegraphics[width=1.0\textwidth]{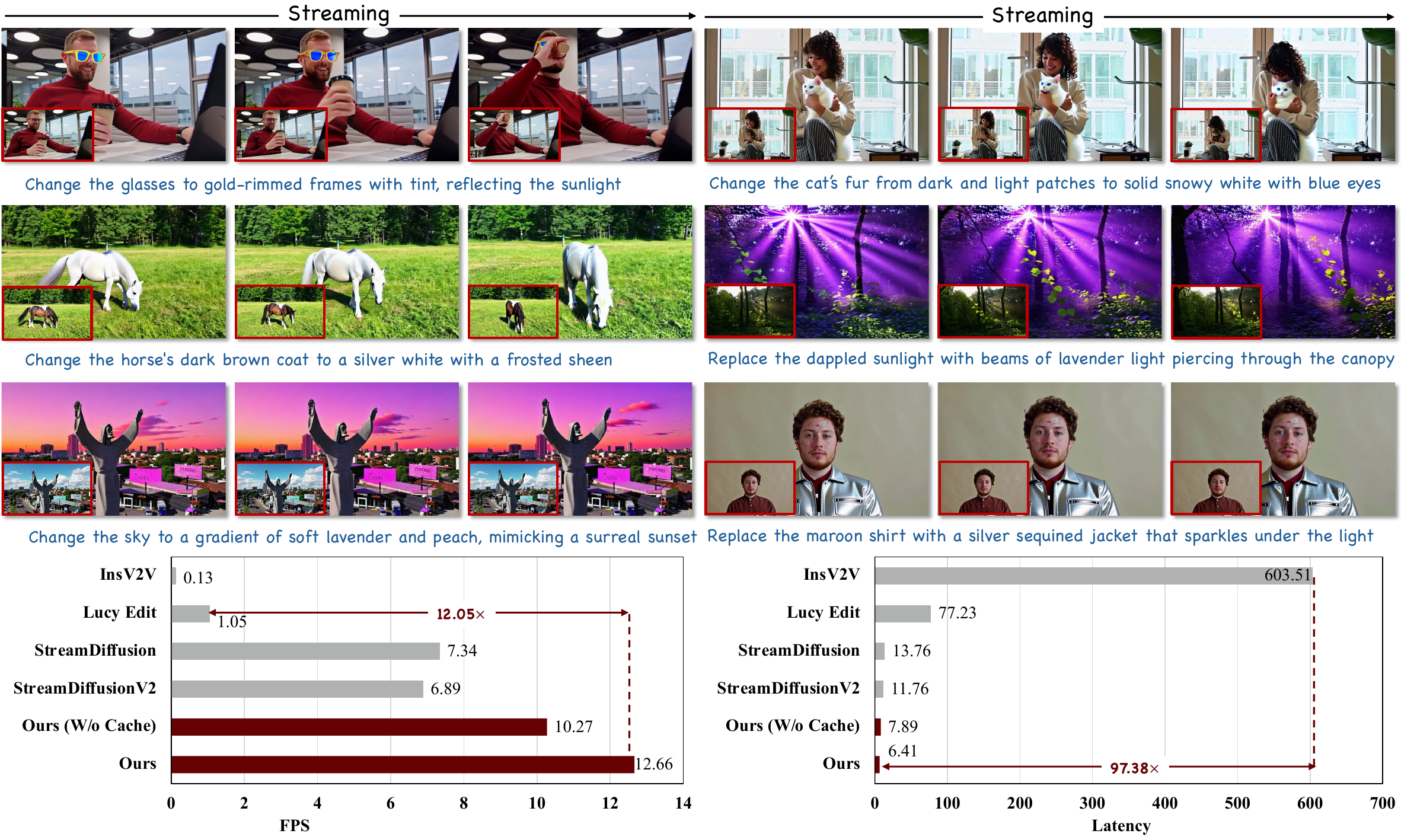}
    \vspace{-0.5em}
    \caption{\textbf{Gallery of various editing results and efficiency comparisons.} We propose the \textit{LiveEdit}, a novel streaming video editing framework capable of performing causal, chunk-by-chunk manipulation with ultra-low latency and strict background preservation. By synergizing a progressive three-stage architectural distillation pipeline with an AR-oriented Mask Cache, \textit{LiveEdit} effectively resolves the architectural incompatibilities and computational bottlenecks inherent in streaming editing paradigms.}
\end{center}
}]

\begingroup
\renewcommand{\thefootnote}{}
\footnotetext{\Letter~Corresponding author.}
\endgroup


\input{sec/0_abstract}

\input{sec/1_introduction}

\input{sec/2_related}

\input{sec/3_method}

\input{sec/4_experiment}
\input{sec/5_conclusion}

\input{sec/X_suppl}
\clearpage  

{
    \small
    \bibliographystyle{ieeenat_fullname}
    \bibliography{main}
}


\end{document}

%% file: preamble.tex
\definecolor{myorange}{RGB}{255,100,3}
\definecolor{mygray}{gray}{.85}
\definecolor{mygray1}{gray}{.7}
\definecolor{mygray2}{gray}{.93}
\definecolor{mygray3}{gray}{.90}
\usepackage{tikz}
\usepackage{array}
\usepackage{graphicx}      
\usepackage{multirow}      
\usepackage{colortbl}      
\usepackage{booktabs}      
\usepackage{xcolor}        
\usepackage{caption}       
\usepackage{gensymb}
\newcolumntype{C}[1]{>{\centering\arraybackslash}p{#1}}

\usepackage{pifont}
\definecolor{iccvblue}{rgb}{0.21,0.49,0.74}
\definecolor{Blue}{RGB}{12, 114, 186} 
\usepackage{color, xcolor}
\usepackage{svg}
\usepackage{pifont}
\usepackage{graphicx}
\usepackage{adjustbox}
\usepackage{algorithm}
\usepackage{algpseudocode}
\usepackage{amssymb}
\usepackage{multirow}
\usepackage{xcolor} 
\usepackage{makecell}
\usepackage{wrapfig}



%% file: sec/0_abstract.tex
\begin{abstract}
Streaming video editing has made rapid progress, yet practical deployment is still limited by two core issues: maintaining stable backgrounds and non-edited regions over time, and achieving the low latency required for real-time interactive scenarios. Meanwhile, recent streaming video generation methods are mostly developed for synthesis and cannot be directly applied to editing due to the strict preservation requirement and region-specific control. In this work, we present a novel streaming video editing framework that performs causal, frame-by-frame editing with strong content preservation and real-time responsiveness. Our key design is a three-stage distillation pipeline that progressively transfers editing capability from a powerful bidirectional foundation model to an efficient unidirectional streaming editor, enabling stable long-horizon edits without sacrificing visual fidelity. To further support real-time deployment, we introduce an AR-oriented mask cache that reuses region-related computation across frames, substantially reducing redundant processing and accelerating inference. Finally, we establish a dedicated benchmark for streaming video editing. Extensive evaluations demonstrate that our method achieves state-of-the-art visual quality among streaming baselines while drastically boosting inference speed to 12.66 FPS, making it suitable for interactive and augmented reality applications.
\end{abstract}

%% file: sec/1_introduction.tex
\section{Introduction}
Video editing~\cite{jiang2025vace,li2025egoedit,decart2025lucyedit,qi2023fatezero,wu2023tune,zhang2023controllable, chen2025contextflow,wang2024taming, feng2025dit4edit, wang2024cove, ma2025magicstick, yang2025unified, zhao2026sana} has witnessed significant advancements, driven by the increasing demand for high-quality content creation and interactive digital experiences. As augmented reality and live-streaming applications become more prevalent, the industry's focus is shifting from traditional offline batch processing toward real-time, responsive editing. However, achieving practical, low-latency deployment remains a formidable challenge, particularly when moving toward a streaming paradigm where video must be processed chunk-by-chunk without access to future information.

\begin{figure}[ht]
    \centering
    \includegraphics[width=1.0\linewidth]{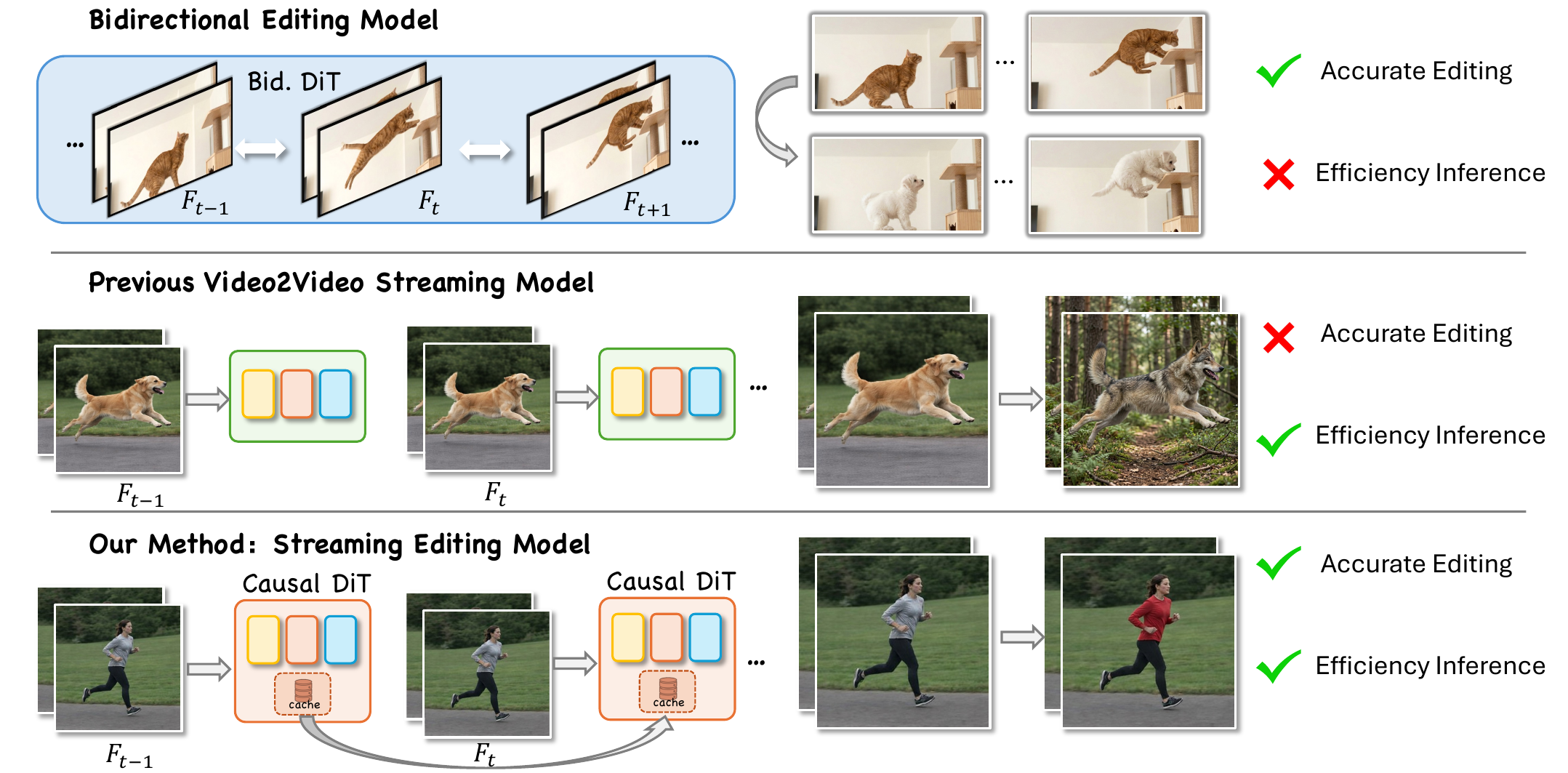}
    \caption{\textbf{Comparison of video editing paradigms.} Unlike bidirectional models that suffer from inefficient inference, and past streaming models that fail to preserve accurate unedited content, our proposed streaming editing model leverages a Causal DiT with a mask-guided cache mechanism to achieve high-fidelity and efficient editing.}
    \label{fig:motivation}
\end{figure}

As illustrated in Fig.~\ref{fig:motivation}, the transition to practical streaming video editing is currently obstructed by two fundamental bottlenecks. {\bf i)} \textbf{Attention distribution shift}: State-of-the-art video diffusion models typically rely on bidirectional or global attention to maintain temporal consistency. Directly adapting these non-causal models to a causal streaming setting—where future frames are unavailable—often leads to a "forgetting" effect or severe flickering, as the model lacks the global structural context required for stable editing. {\bf ii)} \textbf{Spatial-temporal token redundancy}: Standard diffusion pipelines treat every frame as an independent, heavy generation task. However, in autoregressive (AR) oriented streaming generation, the majority of the background remains static or undergoes predictable linear motion. Repeatedly applying dense Feed-Forward Network (FFN) and Attention modules to these redundant, unedited regions leads to prohibitive per-frame latency, making real-time interactive experiences unattainable on edge devices.

To bridge this gap, we introduce a novel framework designed for causal, chunk-by-chunk video editing with strong content preservation and ultra-low latency. Our approach addresses the latency-stability trade-off through a structured three-stage distillation pipeline that progressively transfers the editing capabilities of a powerful bidirectional foundation model to an efficient, unidirectional streaming editor.

Specifically, \textbf{Stage 1 (Foundation Tuning)} focuses on equipping a Bidirectional Diffusion Transformer with robust editing abilities. By leveraging full attention mechanisms and text embeddings, the model learns complex editing mappings supervised by an $\mathcal{L}_{MSE}$ loss. To bridge the gap between offline and streaming processing, \textbf{Stage 2 (Teacher Forcing for Chunk-wise Causal Initial)} transitions the architecture from a bidirectional to a Causal DiT. We employ a teacher-forcing strategy equipped with chunk-wise causal attention, ensuring the model successfully adapts to sequential, unidirectional inputs while strictly maintaining the visual quality and editing priors established in the first stage. Finally, \textbf{Stage 3 (DMD for Streaming Video Editing)} performs advanced distillation to achieve real-time inference. By integrating Distribution Matching Distillation (DMD) with a frozen Real Score and a trainable Fake Score model, we compress the generation process into merely 4 steps. This stage is optimized via both $\mathcal{L}_{MSE}$ and $\nabla_{\theta}\mathcal{L}_{DMD}$ gradients, utilizing pruned noise inputs to further accelerate streaming video editing.

Beyond architectural distillation, redundant background processing remains a critical bottleneck for AR applications. To address this, we introduce an \textbf{AR-oriented Mask Cache} during inference. Instead of running the full DiT forward pass for every frame, our method calculates the $L_2$ distance between the edited output and the source to extract an accurate editing mask. We observe a functional divergence in how different modules handle redundant information: while the Feed-Forward Network and Cross-Attention are essential for maintaining per-pixel spatial detail and text-conditioned alignment, the Self-Attention layers exhibit significant spatio-temporal redundancy in unedited regions. This selective spatial-temporal reuse significantly reduces per-frame latency while guaranteeing that the editing quality remains indistinguishable from the full-calculation baseline.

Our contributions are summarized as follows:

\begin{itemize}

\item We analyze the property of streaming video editing and present the novel streaming video editing framework capable of performing causal, chunk-by-chunk editing with high fidelity and ultra-low inference latency(12.66 FPS).

\item Technically, we first design a comprehensive three-stage distillation pipeline that effectively migrates complex editing knowledge from a Bidirectional DiT teacher to an ultra-efficient, 4-step Causal DiT student. Moreover,  the AR-oriented Mask Cache mechanism is proposed by leveraging $L_2$ distance to dynamically decouple computation.

\item We establish a dedicated benchmark for streaming video editing and demonstrate that our method achieves state-of-the-art performance in terms of both visual quality, temporal consistency, and throughput.

\end{itemize}

%% file: sec/2_related.tex

\section{Related Work}
\label{sec:related_work}

\noindent \textbf{Video Generation and Controllable Editing.}
\label{subsec:video_editing}
Diffusion models have significantly advanced video generation~\cite{singer2022make,blattmann2023stable,yang2024cogvideox, ma2026group, ma2024followpose, ma2025followcreation, ma2026fastvmt, ma2025followyourmotion} and editing. To achieve versatile control~\cite{ma2025controllable}, unified frameworks like VACE~\cite{jiang2025vace} propose all-in-one architectures, while EditVerse~\cite{ju2025editverse} and UNIC~\cite{ye2025unic} employ in-context learning to handle diverse tasks. These approaches typically integrate source videos and dense multi-modal conditions into massive joint representations or extended context sequences. Meanwhile, models like InsV2V~\cite{cheng2023consistent} and Lucy Edit~\cite{decart2025lucyedit} leverage synthetic datasets or channel-wise concatenation for high-fidelity modifications~\cite{mokady2023null}. Additionally, reinforcement learning has been explored to further align generated content with human intents~\cite{black2023training}.

Despite their impressive visual quality, these methods inherently rely on offline, non-causal processing. Processing entire video frames and complex conditions jointly quadratically increases the computational overhead. Consequently, these models must observe the entire temporal context before outputting the initial frame. This unacceptable latency intrinsically hinders their deployment in real-time augmented reality scenarios. To break this paradigm, we propose a novel streaming video editing framework explicitly designed for extreme real-time responsiveness.

\noindent \textbf{Streaming Autoregressive Video Models.}
\label{subsec:streaming_models}
Streaming autoregressive generation effectively overcomes the latency bottlenecks of offline models. Foundational works like Diffusion Forcing~\cite{chen2024diffusion} and StreamDiffusion~\cite{kodaira2025streamdiffusion} redefine denoising as block-wise sequential processing. This paradigm enables massive-scale world models (MAGI-1~\cite{teng2025magi}) and infinite-length cinematic generation (SkyReels-V2~\cite{chen2025skyreels}, StreamingT2V~\cite{henschel2025streamingt2v}). To enhance stability and efficiency, Rolling Forcing~\cite{liu2025rolling} suppresses error accumulation, Stable Video Infinity~\cite{li2025stable} mitigates autoregressive drift for infinite-length generation via error-recycling fine-tuning, Self Forcing~\cite{huang2025self} bridges exposure bias via self-generated conditioning, and StreamDiffusionV2~\cite{feng2025streamdiffusionv2} introduces sink-token-guided rolling KV caches for ultra-low-latency live generation. Furthermore, EgoEdit~\cite{li2025egoedit} pioneered streaming models for egocentric video editing by maintaining temporal consistency in continuous first-person visual translations.

However, directly migrating these generation-tailored mechanisms to general video editing introduces a fundamental misalignment. Video generation synthesizes motion from scratch, heavily relying on past generated predictions (e.g., Self Forcing's serial feedback). Conversely, streaming editing is strongly conditioned on continuous source video streams, prioritizing precise spatial restoration over free-form motion synthesis. Serially depending on historical outputs or employing generation-oriented caches (e.g., DeepCache~\cite{ma2024deepcache}, VMem~\cite{li2025vmem}, or StreamDiffusionV2~\cite{feng2025streamdiffusionv2}) degrades high-frequency structural details when rigorously aligning with semantic editing trajectories. Discarding redundant serial feedback, we introduce an AR-based Cache that achieves extreme computational compression while ensuring strict temporal consistency and spatial alignment.

\noindent \textbf{Efficiency and Distillation in Diffusion.}
\label{subsec:distillation}
Accelerating reverse diffusion~\cite{song2023consistencymodels,wang2023videolcm,salimans2022progressive, zheng2025compute, liu2025survey, zheng2026forecast} is crucial for real-time performance. Early one-step distillation efforts focused on the image domain, utilizing techniques like Rectified Flow (InstaFlow~\cite{liu2023instaflow}), target score distillation (TSD-SR~\cite{Dong_2025_CVPR}), and Distribution Matching Distillation~\cite{yin2024improved,yin2025slow}, alongside progressive adversarial distillation (SDXL-Lightning~\cite{lin2024sdxl}). These advancements have naturally extended to the video domain, enabling efficient one-step generation (AAPT~\cite{lin2025diffusion}), restoration (SeedVR2~\cite{wang2025seedvr2}), and super-resolution (DOVE~\cite{chen2025doveefficientonestepdiffusion}). Sparse VideoGen2~\cite{yang2025sparsevideogen2acceleratevideo} further reduces redundancy via attention permutation. Orthogonal to distillation, structural optimizations like Token Merging~\cite{bolya2022token} and FlashAttention~\cite{dao2022flashattention} are widely adopted to alleviate system memory overheads.

Most closely related to our work is FlashVSR~\cite{zhuang2025flashvsr}, which achieves real-time streaming video super-resolution via a three-stage distillation pipeline designed to progressively accelerate latent rendering. Similarly, PersonaLive~\cite{li2025personalive} leverages appearance distillation for live portrait animation. However, these tasks fundamentally involve low-level pixel mapping or structurally constrained regions. Directly transferring these distillation schemes to general video editing—which involves complex semantic reconstruction, cross-scale feature evolution, and training-free reward-guided control—frequently degrades high-fidelity details. In contrast, we propose a tailored three-stage distillation strategy integrated with our AR-based Cache, achieving unprecedented inference acceleration without sacrificing complex editing fidelity.

%% file: sec/3_method.tex
\section{Method}
\label{sec:method}

\subsection{Motivation}

\begin{figure}[ht]
    \centering
    \includegraphics[width=1.0\linewidth]{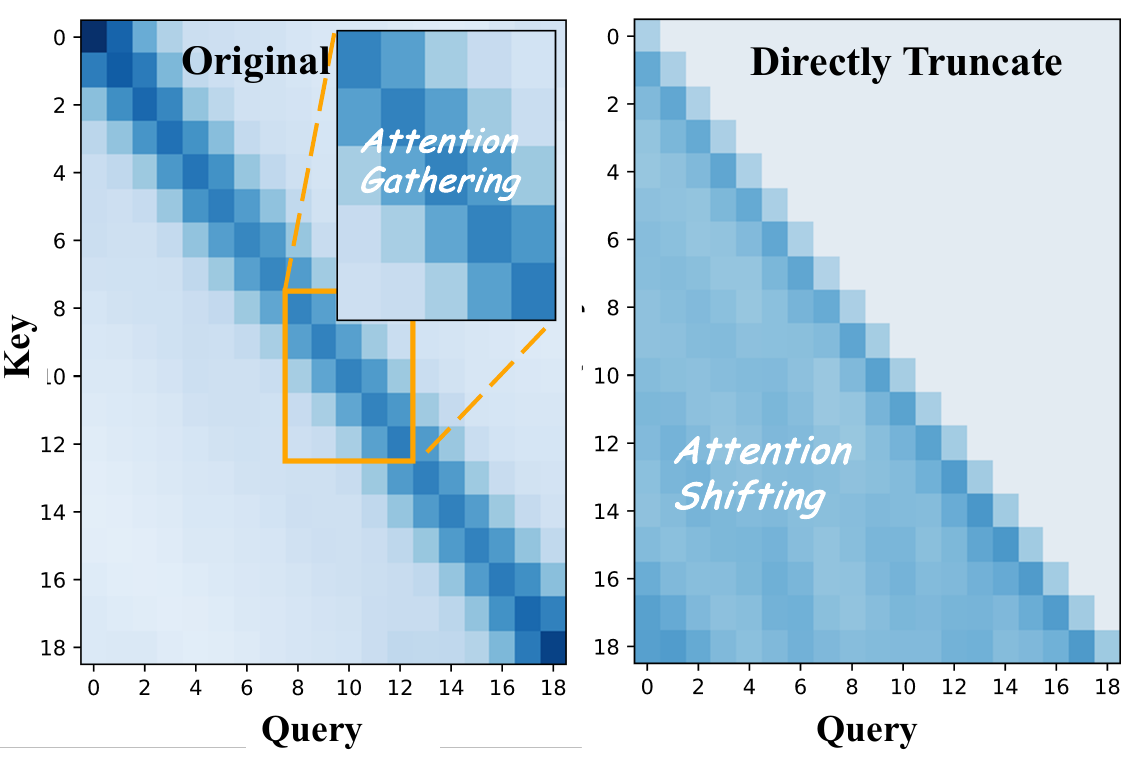}
    \caption{\textbf{Visualization of the attention distribution shift.} \textbf{Left:} The bidirectional prior exhibits localized attention gathering. \textbf{Right:} Direct causal truncation forces attention to spread uniformly across all historical frames.}
    \label{fig:attention_shift}
\end{figure}

We summarize the two primary observations regarding the inefficiencies of adapting state-of-the-art video diffusion models to the streaming video editing task and propose the modules to address them.

\noindent \textbf{Attention distribution shift.} In the offline video editing phase, state-of-the-art bidirectional diffusion models utilize dense temporal attention to propagate structural information, natively assigning significant weights to both past and immediate future tokens. However, we observe that abruptly truncating these future keys and values for causal execution causes a severe shift in the attention distribution. Specifically, as illustrated in Fig.~\ref{fig:attention_shift}, the loss of future context forces the attention weights to flatten and distribute uniformly across all available historical frames. This homogenized attention behavior intrinsically conflicts with the streaming paradigm, which relies heavily on the nearest neighboring frames to maintain temporal coherence and structural integrity. Consequently, this over-smoothed dependency disrupts the pre-trained structural priors, making it theoretically suboptimal to directly deploy naively truncated bidirectional models in a streaming pipeline. To address this, we introduce a progressive three-stage distillation pipeline. Our teacher-forcing mechanism explicitly aligns the causal attention distribution with the localized bidirectional prior, ensuring robust, localized representation mapping without the need for future context.

\noindent \textbf{Spatial-temporal token redundancy.} Existing streaming generation approaches (e.g., StreamV2V~\cite{liang2024looking}, StreamDiffusion~\cite{kodaira2025streamdiffusion} and StreamDiffusionV2~\cite{feng2025streamdiffusionv2}) primarily focus on global Video-to-Video translation tasks. Specifically, for every incoming frame, these pipelines perform dense computations across all spatial tokens globally to synthesize the entire scene. However, we note that streaming video editing fundamentally differs from global translation, as the intermediate feature representations of tokens in unedited regions must strictly maintain absolute temporal consistency. Therefore, applying such a global generation paradigm to editing tasks inherently disrupts the visual integrity of unedited areas, causing destructive structural degradation and severe flickering in the background, alongside massive computational redundancy. To address this, we introduce the AR-oriented Mask Cache mechanism. Only the actively edited spatial tokens undergo full calculation, while the intermediate features of highly correlated static tokens are directly retrieved from the cache, ensuring strict background preservation and efficient generation.

\subsection{Three-Stage Distillation Pipeline}

\begin{figure*}[ht]
    \centering
    \includegraphics[width=1.0\linewidth]{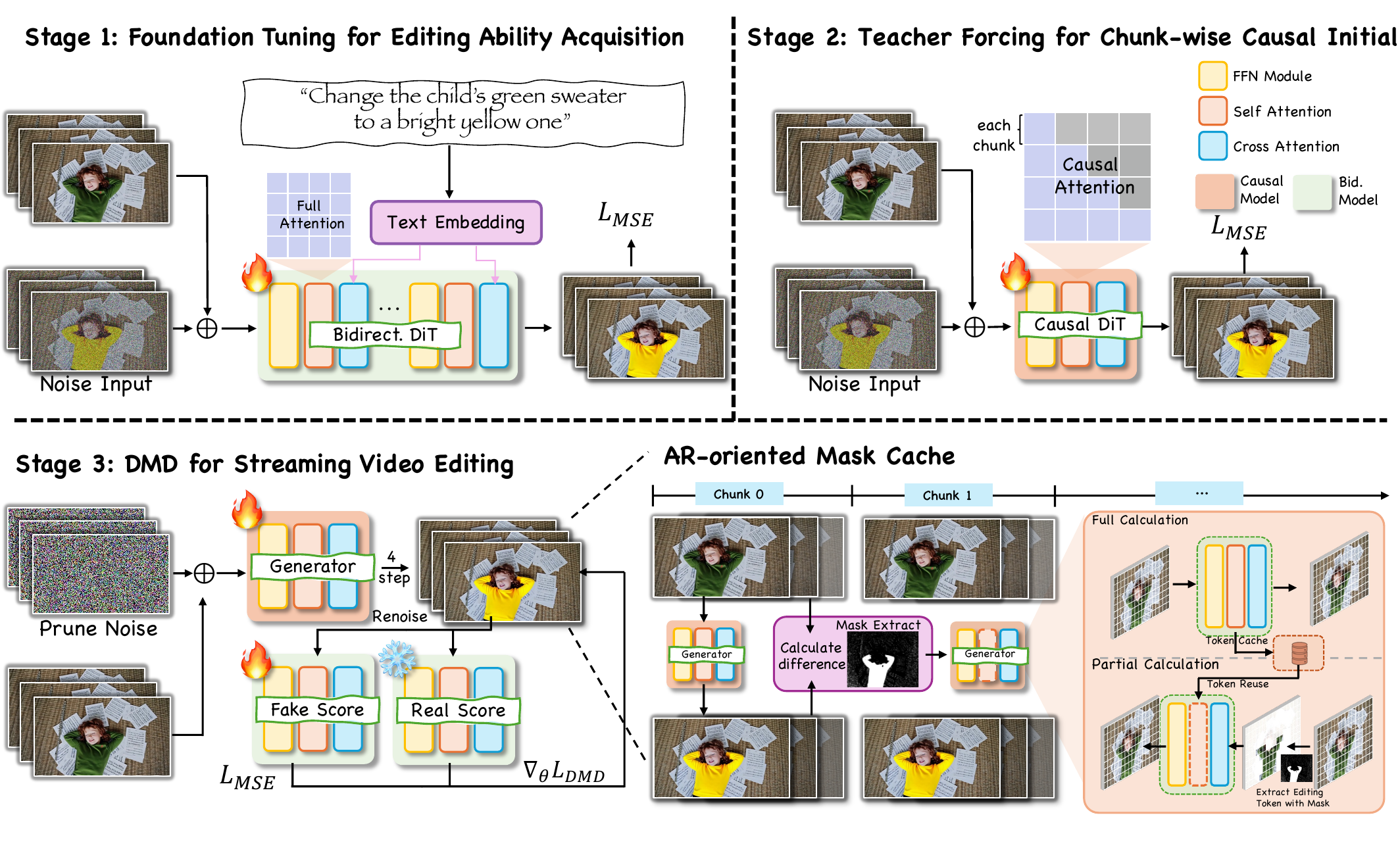}
    \vspace{-2em}
    
    \caption{\textbf{Overview of the proposed streaming video editing framework.} Our approach features a three-stage distillation pipeline that transfers editing capabilities from a bidirectional DiT to a 4-step causal model. Furthermore, an AR-oriented Mask Cache accelerates real-time inference by dynamically decoupling computation and reusing tokens in unedited background regions.}
    
    \label{fig:framework}
\end{figure*}

To bridge the architectural gap between offline bidirectional priors and online causal execution, we propose a progressive three-stage distillation pipeline. Let $z_0\in\mathbb{R}^{F\times C\times H\times W}$ denote the input video latent sequence and $c$ represent the text embedding extracted from the editing prompt. This design effectively transfers the high-fidelity editing capabilities of a foundation model into an ultra-fast, unidirectional streaming editor. An overview of the framework is illustrated in Fig. \ref{fig:framework}.

\noindent \textbf{Stage 1: Foundation Tuning for Editing Ability Acquisition.}
In the initial stage, we establish a robust multimodal video-to-video editing baseline utilizing a Bidirectional Diffusion Transformer (DiT). Specifically, the network processes the channel-wise concatenation of the original video latent and the noisy latent $z_t$ at timestep $t$. To mitigate the quadratic computational overhead inherently associated with long token sequences in video generation, we emphasize this channel-wise integration strategy over spatial or temporal sequence concatenation. The bidirectional architecture, denoted as $\epsilon_{\theta}^{bid}$, leverages full temporal and spatial attention to learn complex mapping functions for content manipulation. The entire foundation model is supervised via the standard noise-matching objective:
$$\mathcal{L}_{MSE}^{bid}=\mathbb{E}_{z_0,\epsilon\sim\mathcal{N}(0,I),t,c}\left[\left\|\epsilon-\epsilon_{\theta}^{bid}(z_t,t,c)\right\|_2^2\right]$$
This optimization yields a powerful offline editing prior capable of maintaining high-fidelity generation.

\noindent \textbf{Stage 2: Teacher Forcing for Chunk-wise Causal Initial.}
To enable basic streaming input-output functionality, the architecture must transition from a bidirectional to an autoregressive (AR) generation paradigm. However, directly applying causal masks to a pre-trained bidirectional model severely degrades performance. Therefore, we introduce a Teacher Forcing mechanism equipped with chunk-wise causal attention. Let $M_{causal}$ denote the causal attention mask that restricts temporal tokens from attending to future chunks. The causal DiT, $\epsilon_{\theta}^{causal}$, is optimized to predict the noise while strictly adhering to the causal constraint:
$$\mathcal{L}_{MSE}^{causal}=\mathbb{E}_{z_0,\epsilon,t,c}\left[\left\|\epsilon-\epsilon_{\theta}^{causal}(z_t,t,c\mid M_{causal})\right\|_2^2\right]$$
Inspired by recent autoregressive frameworks such as Causal Forcing, we establish that deriving an AR model via explicit Teacher Forcing is structurally essential for streaming video generation. By aligning the output distribution of the causal DiT with the bidirectional representations learned in Stage 1, the model prevents the structural collapse typically caused by the absence of future tokens.

\noindent \textbf{Stage 3: DMD for Streaming Video Editing.}
To achieve ultra-low latency while eliminating accumulated shift errors during continuous streaming, we perform advanced step-distillation using Distribution Matching Distillation (DMD). Recent autoregressive generation paradigms, notably Self-Forcing, heavily rely on an Ordinary Differential Equation (ODE) initialization phase to bridge the training-inference distribution gap. However, this ODE initialization incurs prohibitive computational resource overhead and severely limits practical scalability. To circumvent this fundamental bottleneck, we directly utilize the AR-based model parameters from Stage 2 ($\epsilon_{\theta}^{causal}$) to initialize the 4-step DMD generator $G_{\theta}$. This architectural decision not only avoids the excessive initialization overhead but also provides a highly stable starting point for distillation, echoing the empirical insights observed in recent works like EgoEdit.

During training, the generator $G_{\theta}$ maps pruned noise inputs directly to the edited frames. The distillation process is jointly optimized by $\mathcal{L}_{MSE}$ and the DMD gradient $\nabla_{\theta}\mathcal{L}_{DMD}$. The DMD gradient is computed between a frozen Real Score model $\epsilon_{\phi}^{real}$ and a trainable Fake Score model $\epsilon_{\psi}^{fake}$:

\begin{align*}
\nabla_{\theta}\mathcal{L}_{DMD}
&=\mathbb{E}_{z_T,c}\Bigl[
w(t)\Bigl(\epsilon_{\phi}^{real}(z_t,t,c)- \\
&\qquad
\epsilon_{\psi}^{fake}(z_t,t,c)\Bigr)
\nabla_{\theta}G_{\theta}(z_T,c)\Bigr]
\end{align*}

where $z_t$ is the intermediate latent simulated from the generated output $G_{\theta}(z_T,c)$, and $w(t)$ is a timestep-dependent weighting function. This mechanism effectively compresses the generation process into merely 4 inference steps, ensuring real-time responsiveness.

\subsection{AR-oriented Mask Cache}

\begin{figure*}[ht]
    \centering
    \includegraphics[width=1.0\linewidth]{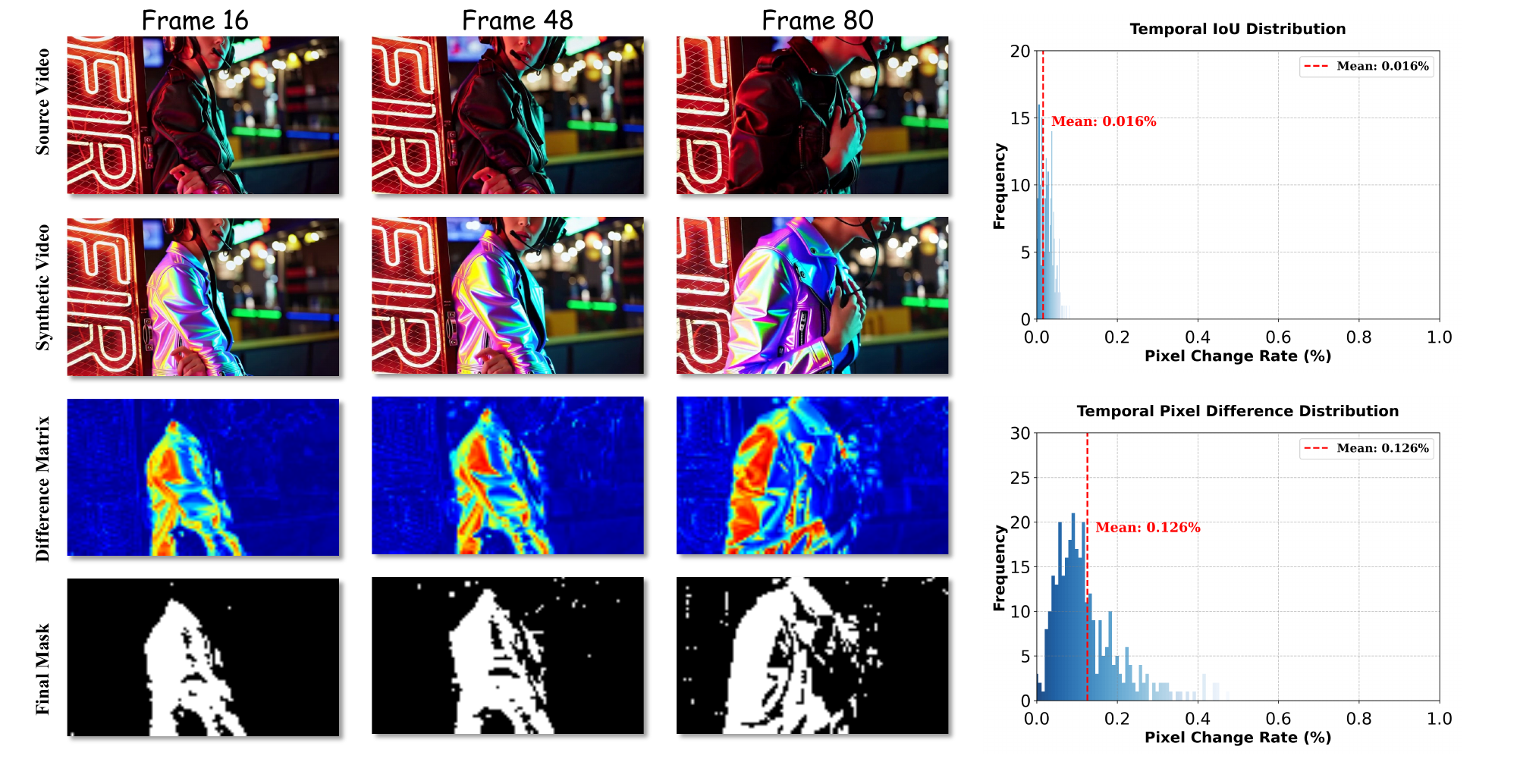}
    \caption{\textbf{Visualization of the temporal consistency analysis and mask generation process.} The left panels show (from top to bottom) the source video frames, the synthesized video frames, the computed difference matrices, and the resulting binary masks. The right panels display the statistical distributions of Temporal IoU and Pixel Difference across the sequence, with mean values of 0.016\% and 0.126\%, respectively, indicating high structural stability.}
    \label{fig:mask}
\end{figure*}

While the three-stage distillation pipeline resolves the architectural incompatibility, the fundamental issue of spatial-temporal redundancy remains a critical bottleneck for real-time inference. To address this, we introduce the AR-oriented Mask Cache mechanism, dynamically decoupling the computational graph based on regional editing activity.

During the streaming inference phase, the pipeline processes the video in sequential chunks. To dynamically route the computation for an incoming chunk $k$, we derive its spatial editing mask from the generation trajectory of the preceding chunk. Let $z_{src}^{k-1}$ denote the original source latent representation of the previously generated chunk $k-1$, and $z_{edit}^{k-1}$ denote its corresponding edited output latent. We extract a binary spatial editing mask $M^k\in\{0,1\}^{H\times W}$ for the current chunk $k$ by computing the $L_2$ distance between these two latents:

$$M^k_{u,v}=\mathbb{I}\left(\left\|z_{edit,u,v}^{k-1}-z_{src,u,v}^{k-1}\right\|_2>\tau\right)$$

where $\mathbb{I}(\cdot)$ is the indicator function and $\tau$ is dynamically determined by calculating a redundancy level among the tokens. This mask geometrically separates the spatial layout of the incoming chunk $k$ into active editing regions ($M^k_{u,v}=1$) and static background regions ($M^k_{u,v}=0$). As visualized in Fig. \ref{fig:mask}, our mask generation process demonstrate extremely high structural stability, with the statistical distribution across the entire sequence.

Instead of executing the complete network forward pass globally for chunk $k$, our mechanism implements a spatial routing strategy guided by $M^k$. For the spatial tokens located within the active mask, the model allocates its full computational capacity, executing the complete sequence of Self-Attention, Cross-Attention, and Feed-Forward Network modules (\textit{Full Calculation}). Conversely, for the tokens corresponding to the unedited regions, the mechanism entirely bypasses the computationally expensive layers. Let $\mathcal{F}$ denote the full block transformation and $z^k_{u,v}$ denote the input token for the current chunk; the output token feature $f^k_{u,v}$ is determined by:

$$f^k_{u,v}=\begin{cases}\mathcal{F}(z^k_{u,v})&\text{if }M^k_{u,v}=1\\f^{k-1}_{u,v}&\text{if }M^k_{u,v}=0\end{cases}$$

The intermediate feature representations for the static regions are directly retrieved from a maintained Token Cache populated by the preceding chunk (\textit{Token Reuse}). This dynamic, inter-chunk spatial decoupling dramatically reduces the per-frame computational complexity, achieving substantial acceleration while strictly guaranteeing absolute visual consistency in the unedited background areas.

%% file: sec/4_experiment.tex
\begin{figure*}[t]
    \centering
    \includegraphics[width=1.0\linewidth]{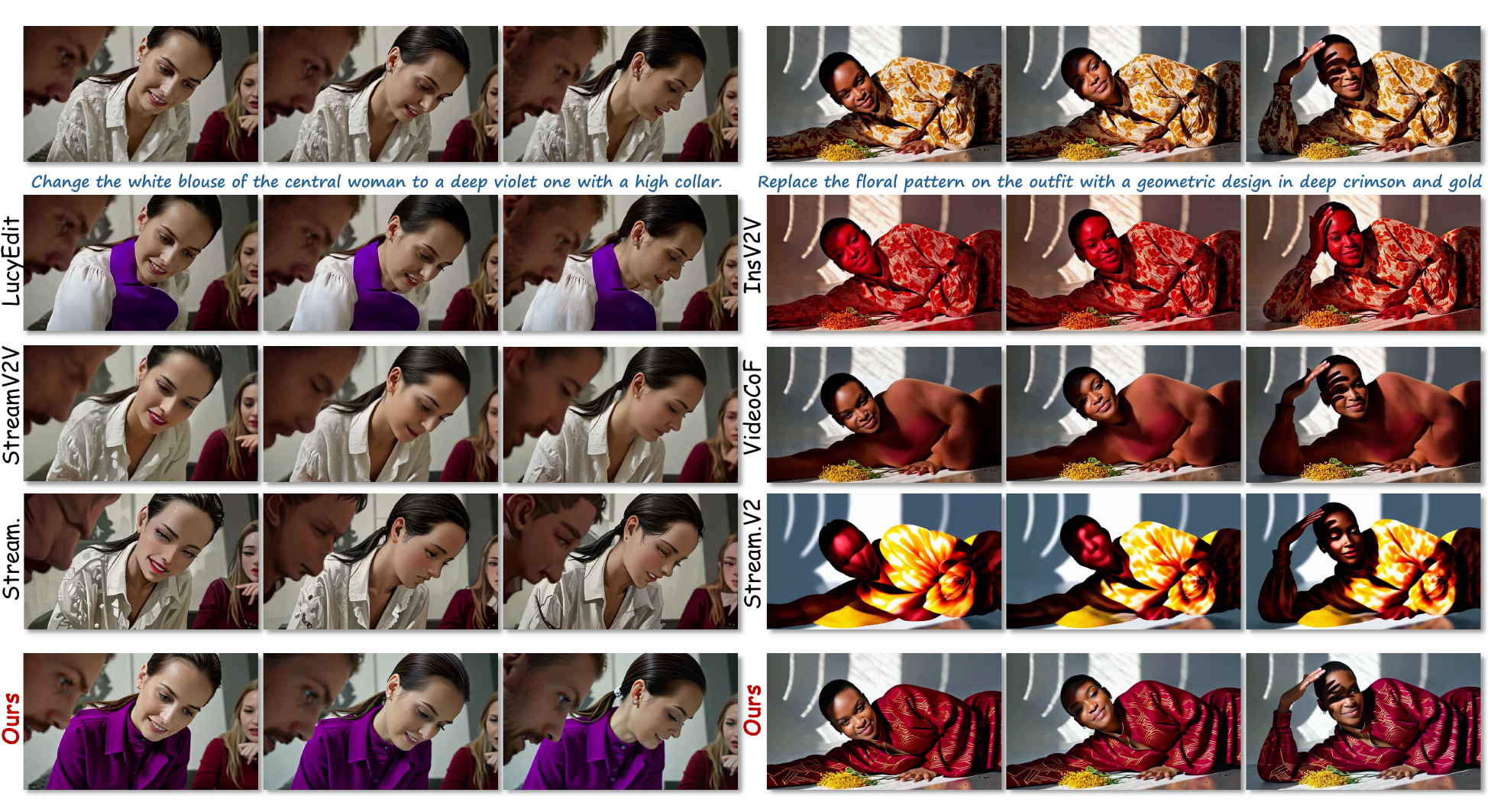}
    \caption{\textbf{Qualitative comparison of streaming video editing performance.} The source videos and instructions are displayed at the top. While existing methods exhibit significant limitations, leading to structural collapse or an inability to accurately follow the text, our approach precisely modifies the target regions and preserves the visual quality and temporal coherence of the original scenes.}
    \label{fig:compare}
\end{figure*}

\section{Experiment}
\label{sec:experiment}

\subsection{Implementation Details}

\noindent \textbf{Model Configuration.} We build our foundation model upon Wan2.1-T2V-1.3B~\cite{wan2025wanopenadvancedlargescale}. During Stage 1, we explicitly employ channel-wise concatenation to integrate the noisy latent $z_t$ and the condition latent, rather than appending them token-wise along the sequence dimension. By strictly maintaining the original sequence length, this structural design fundamentally bypasses the quadratic computational explosion in attention mechanisms typically associated with spatial-temporal video generation. For training, our dataset comprises 20K high-quality video-video pairs, which are carefully filtered from the large-scale Ditto-1M~\cite{bai2025ditto} dataset.

\noindent \textbf{Training Setup.} The three-stage distillation pipeline is trained progressively on 8 NVIDIA A100 GPUs. We employ the AdamW optimizer across all stages. 
\begin{itemize}
    \item \textbf{Stage 1 (Foundation Tuning):} The bidirectional foundation model $\epsilon_{\theta}^{bid}$ is trained for 9K steps with a learning rate of $10^{-5}$ and a global batch size of 8. We utilize a standard noise scheduler with continuous timesteps $t \in [0, 1000]$.
    \item \textbf{Stage 2 (Teacher Forcing):} We transition the bid. Dit to the causal DiT $\epsilon_{\theta}^{causal}$ by introducing the chunk-wise causal attention mask $M_{causal}$. The model is fine-tuned for an additional 20K steps. The temporal chunk size is set to 3 latent frames, ensuring that tokens can only attend to the current and strictly preceding chunks.
    \item \textbf{Stage 3 (DMD):} The 4-step generator $G_{\theta}$ is initialized directly from the Stage 2 weights. deliberately bypassing the computationally expensive ODE initialization. This bypass is explicitly enabled by the autoregressive training conducted in Stage 2, which provides a well-aligned causal distribution starting point. For the 4-step generation, the sampling timesteps are specifically set to $[0, 250, 500, 750]$. Both the Real Score model and the Fake Score model are initialized from the foundation model weights obtained in Stage 1. We apply a reduced learning rate of $10^{-5}$ for 10K steps, utilizing the timestep-dependent weighting function $w(t)$ as proposed in standard DMD formulations.

\end{itemize}

\noindent \textbf{Inference and Mask Cache.} During the streaming inference phase, the model operates in a purely autoregressive manner, decoding 3 frames per step. For the AR-oriented Mask Cache, rather than using a fixed empirical value, the $L_2$ distance threshold $\tau$ for the binary spatial editing mask $M^k$ is dynamically calculated to explicitly prune $70\%$ of the redundant spatial tokens. Specifically, this caching mechanism is applied within the Self-Attention layers, as we empirically find that applying the cache to the Self-Attention computation introduces no degradation in generation quality. Tokens in these pruned regions bypass the standard attention computation, and their intermediate features are directly retrieved from the Token Cache. This dynamic routing allows the 4-step streaming editing to achieve an ultra-low latency of $79ms$ per-frame, ensuring real-time responsiveness without background flickering.

\subsection{Qualitative comparison.}

\label{subsec:qualitative_comparison}

We present the visual comparison between our proposed framework and two distinct categories of state-of-the-art video editing baselines: bidirectional offline editing models (LucyEdit~\cite{decart2025lucyedit}, InsV2V~\cite{cheng2023consistent}, VideoCoF~\cite{yang2025unified}) and streaming generation models (StreamV2V~\cite{liang2024looking}, StreamDiffusion~\cite{kodaira2025streamdiffusion}, StreamDiffusionV2~\cite{feng2025streamdiffusionv2}). As illustrated in Fig. \ref{fig:compare}, existing methods exhibit significant limitations in balancing precise prompt alignment with structural preservation.

StreamV2V fundamentally struggles to execute localized, precise attribute modifications, failing to apply the requested edits and leaving the source attributes largely unaltered. Conversely, approaches such as InsV2V and VideoCoF suffer from severe color bleeding because they indiscriminately apply target attributes across the entire frame, thereby corrupting the subject's skin tone and the surrounding environment. Furthermore, methods relying on global translation paradigms inherently compromise the visual integrity of unedited areas; StreamDiffusion and StreamDiffusionV2 exhibit severe structural collapse and unintended style shifts, whereas LucyEdit lacks the strict spatial control necessary to preserve delicate structural details. Overcoming these limitations, our method accurately interprets complex textual conditions to apply distinct textures and colors exclusively to target regions. Benefiting from the proposed AR-oriented Mask Cache mechanism, our approach explicitly decouples the generation process, guaranteeing zero visual degradation in unedited background regions. Consequently, complex lighting conditions, shadows, and subject identities are strictly preserved alongside high-fidelity edits.

\begin{figure*}[t]
    \centering
    \includegraphics[width=1.0\linewidth]{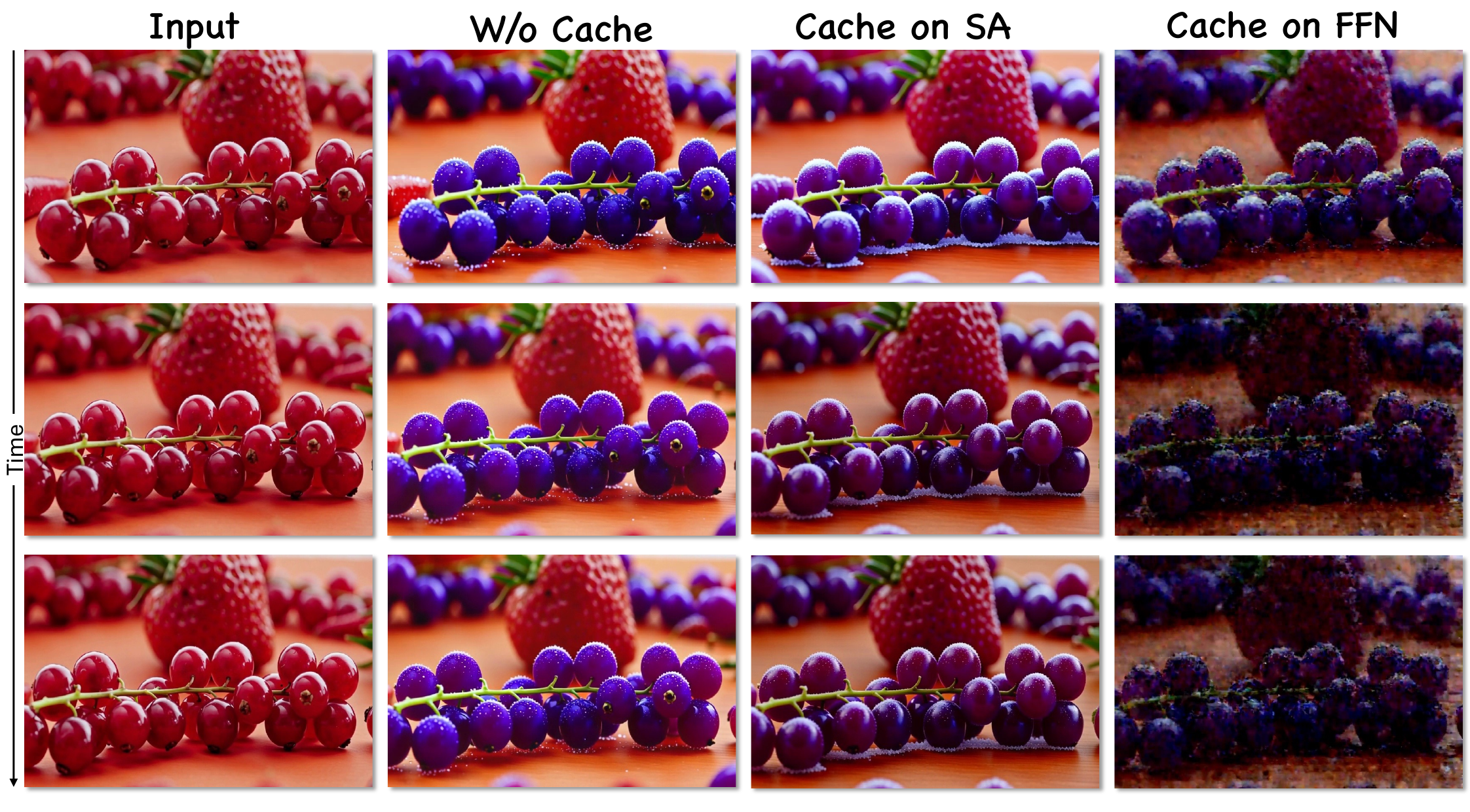}
    \caption{\textbf{Visual comparison of different cache locations.} With the instruction \textit{"Change the red currants to deep purple grapes with a thin layer of frost on their skins"}.}
    \label{fig:cache_layer}
\end{figure*}

\subsection{Quantitative comparison.}
\input{tab/baseline}
We compare our method against the state-of-the-art baselines on a collected benchmark consisting of 120 pairs. To comprehensively assess both the generation quality and editing accuracy, we employ six standard automated metrics, with the results summarized in Tab.~\ref{tab:comparison_videoediting_metrics}. 

For Text Alignment, following EgoEdit~\cite{li2025egoedit}, we compute the CLIP~\cite{radford2021learning} feature similarity of the edited results with the text prompt to evaluate editing consistency and instruction adherence. To assess the aesthetic appeal, we utilize the LAION-Aesthetic Score Predictor~\cite{laion2022aesthetics} for the Aesthetic Quality metric. Furthermore, we employ the VBench evaluation framework~\cite{huang2024vbench} to measure Background Consistency, Motion Smoothness, Dynamic Degree, and Imaging Quality, which collectively provide a comprehensive assessment of the overall visual fidelity.

As demonstrated in Tab.~\ref{tab:comparison_videoediting_metrics}, our proposed framework achieves best performance across almost all dimensions. Notably, while bidirectional offline architectures naturally benefit from future temporal context, our unidirectional streaming approach not only bridges the performance gap but strictly outperforms them in Text Alignment (achieving 0.270 compared to InsV2V's 0.259). Furthermore, our method achieves the highest Dynamic Degree and Imaging Quality among all evaluated methods, while maintaining highly competitive Aesthetic Quality against strong streaming baselines like StreamDiffusion. Crucially, the results demonstrate that the proposed AR-oriented cache improves Text Alignment and Motion Smoothness while perfectly preserving Background Consistency, proving that our mechanism effectively maintains high visual integrity during continuous sequential processing. Furthermore, we invited 20 volunteers to rank methods across background consistency, editing fidelity, and overall quality. The results are provided in the supplementary materials.

\input{tab/ablation-arch}

\subsection{Ablation Study}
\label{subsec:ablation_study}

\input{tab/ablation_cache}

We conduct an ablation study to validate the necessity of each phase within our three-stage distillation pipeline, with the configuration progression summarized in Tab.~\ref{tab:three_stages_ablation}.

\noindent \textbf{Effectiveness of Stage 1 (Foundation Tuning).} The initial phase establishes a robust bidirectional prior essential for high-fidelity structural preservation and text alignment. Operating globally across the full sequence, this stage successfully acquires complex editing capabilities. However, it requires 100 Network Function Evaluations (NFEs) and relies on Classifier-Free Guidance (CFG), which strictly prevents online, continuous streaming inference. 

\noindent \textbf{Effectiveness of Stage 2 (Teacher Forcing).} To transition the offline foundation model to a streaming paradigm, Stage 2 introduces chunk-wise causal attention. This structural modification successfully enables autoregressive execution, processing the video continuously in sequential chunks of 3 frames. While this stage achieves the fundamental streaming input-output format, it still requires 100 NFEs and maintains the CFG dependency, resulting in substantial computational latency that prohibits real-time deployment. 

\noindent \textbf{Effectiveness of Stage 3 (DMD).} The final distillation phase dramatically accelerates the inference speed. By compressing the generation process down to 4 NFEs and explicitly eliminating the need for CFG (which otherwise doubles the required forward passes), the computational overhead is drastically reduced. Furthermore, initializing the student generator directly from the autoregressive weights optimized in Stage 2 bypasses the expensive standard ODE initialization. Ultimately, the integration of all three stages yields a framework that simultaneously achieves an ultra-low latency of 7.89s for 81 frames, pure streaming functionality, and high-quality editing performance.

\begin{figure}
    \centering
    \includegraphics[width=\linewidth]{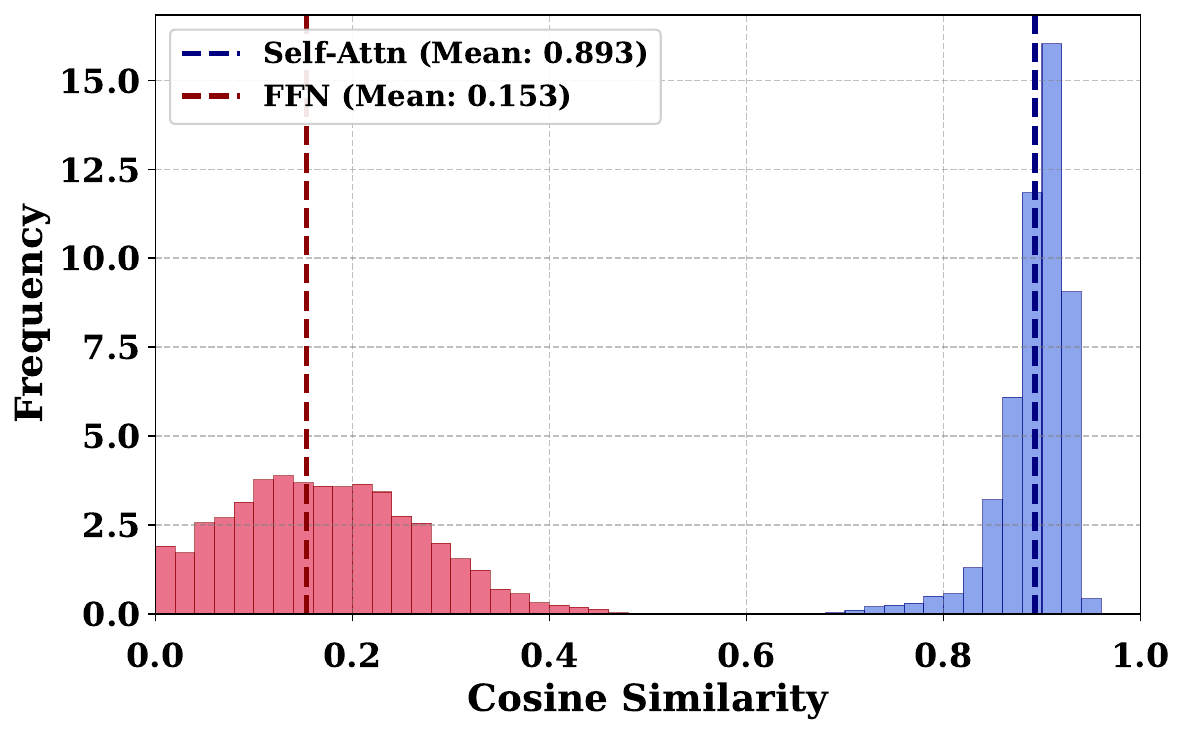}
    \caption{Distribution of token cosine similarity between consecutive denoising step.}
    \label{fig:cosine}
\end{figure}
\noindent\textbf{Effectiveness of AR-oriented Mask Cache.}
To identify the optimal integration point for the AR-oriented Mask Cache, we investigate the impact of applying the caching mechanism to different architectural components, specifically comparing its placement in the Self-Attention (SA) layers versus the FFN modules.

As summarized in Tab.~\ref{tab:ablation_cache}, applying the cache to the SA layers (Ours) achieves superior performance across all evaluated dimensions. In contrast, caching FFN features causes significant degradation, suggesting that FFN layers contain high-frequency spatial information too sensitive for direct temporal reuse. This architectural divergence is explicitly validated by the feature similarity distributions in Fig.~\ref{fig:cosine}, where SA tokens maintain exceptionally high temporal redundancy across consecutive steps, whereas FFN representations exhibit notably low similarity.

The visual evidence in Fig.~\ref{fig:cache_layer} further corroborates these findings. Both the baseline configuration (W/o Cache) and the SA-cache version successfully modify the target object with high fidelity, correctly rendering fine-grained textures and maintaining natural color saturation. Conversely, caching on FFN results in severe blurring and structural instability. These results confirm that caching SA features is the optimal strategy, as it effectively leverages spatial-temporal redundancy to achieve significant acceleration without sacrificing the generative capacity and high-quality spatial details of the full-calculation baseline.

%% file: tab/baseline.tex
\begin{table}[t]
  \centering
  \caption{\textbf{Quantitative comparison of video editing methods.} We evaluate across six metrics: Text Alignment (TA), Background Consistency (BC), Motion Smoothness (MS), Dynamic Degree (DD), Aesthetic Quality (AQ), and Imaging Quality (IQ). \textcolor{red}{\textbf{Red}} and \textcolor{blue}{\textbf{Blue}} denote the best and second-best results, respectively.}
  \label{tab:comparison_videoediting_metrics}
  \addtolength{\tabcolsep}{4pt}
    \resizebox{1.0\linewidth}{!}{%
      \begin{tabular}{lcccccc}
        \toprule
        Method & TA$\uparrow$ & BC$\uparrow$ & MS$\uparrow$ & DD$\uparrow$ & AQ$\uparrow$ & IQ$\uparrow$ \\
        \midrule
        LucyEdit & 0.253 & 0.943 & 0.990 & 0.266 & 0.529 & 0.707 \\
        VideoCoF & 0.245 & 0.953 & 0.991 & 0.094 & 0.542 & 0.709 \\
        InsV2V   & 0.259 & 0.943 & 0.986 & 0.196 & 0.577 & 0.708 \\
        \midrule
        StreamDiffusion   & 0.239 & 0.886 & 0.975 & 0.239 & \textcolor{red}{\textbf{0.590}} & \textcolor{blue}{\textbf{0.717}} \\
        StreamDiffusionV2 & 0.252 & 0.951 & \textcolor{blue}{\textbf{0.992}} & 0.264 & 0.539 & 0.653 \\
        StreamV2V         & 0.244 & 0.934 & 0.989 & 0.153 & 0.548 & 0.712 \\
        \midrule
        \textbf{Ours (W/o Cache)} & \textcolor{blue}{\textbf{0.265}} & \textcolor{red}{\textbf{0.956}} & 0.991 & \textcolor{red}{\textbf{0.282}} & \textcolor{blue}{\textbf{0.584}} & \textcolor{red}{\textbf{0.720}} \\
        \textbf{Ours (W/ Cache)}  & \textcolor{red}{\textbf{0.270}} & \textcolor{blue}{\textbf{0.956}} & \textcolor{red}{\textbf{0.992}} & \textcolor{blue}{\textbf{0.256}} & 0.581 & 0.708 \\
        \bottomrule
      \end{tabular}
    }
\end{table}

%% file: tab/ablation-arch.tex
\begin{table}[t]
  \centering
  \caption{\textbf{Ablation of our three-stage distillation pipeline.} Comparison of generation configurations and inference efficiency across the different training stages.}
    \resizebox{1.0\linewidth}{!}{%
      \begin{tabular}{lccc}
        \toprule
        & \textbf{Stage 1 (Foundation)} & \textbf{Stage 2 (Teacher Forcing)} & \textbf{Stage 3 (DMD)} \\
        \midrule
        Is streaming? & $\times$ & $\checkmark$ & $\checkmark$ \\
        NFEs & 100 & 100 & 4 \\
        With CFG? &  $\checkmark$ & $\checkmark$ & $\times$  \\
        Latency & 197.48 & 200.36 & 7.89 \\
        First chunk size & Full sequence & 3 frames & 3 frames \\
        Next chunk size & N/A & 3 frames & 3 frames \\
        Text Alignment & 0.268 & 0.264 & 0.265 \\
        Image Quality  & 0.716 & 0.702 & 0.720 \\
        \bottomrule
      \end{tabular}
    }
    \vspace{-1em}
  \label{tab:three_stages_ablation}
\end{table}

%% file: tab/ablation_cache.tex
\begin{table}[t]
    \centering
    \caption{\textbf{Quantitative results of the cache-mechanism ablation.} Comparing the performance of applying the caching mechanism to either Self-Attention or FFN layers. \textcolor{red}{\textbf{Red}} denote the best results.}
    \label{tab:ablation_cache}
    \resizebox{0.5\textwidth}{!}{
        \begin{tabular}{lcccccc}
        \toprule
        Method & TA$\uparrow$ & BC$\uparrow$ & MS$\uparrow$ & DD$\uparrow$ & AQ$\uparrow$ & IQ$\uparrow$ \\
        \midrule
        \textbf{W/o Cache}      & 0.265 & \textcolor{red}{\textbf{0.956}} & 0.991 & \textcolor{red}{\textbf{0.282}} & \textcolor{red}{\textbf{0.584}} & \textcolor{red}{\textbf{0.720}} \\
        \textbf{Cache on SA}   & \textcolor{red}{\textbf{0.270}} & 0.956 & \textcolor{red}{\textbf{0.992}} & 0.256 & 0.581 & 0.708 \\
        \textbf{Cache on FFN}  & 0.236 & 0.841 & 0.982 & 0.017 & 0.440 & 0.513 \\
        \bottomrule
      \end{tabular}
    }
\end{table}

%% file: sec/5_conclusion.tex
\section{Conclusion}
\label{sec:conclusion}

 In this paper, we presented LiveEdit, a novel streaming video editing framework that successfully adapts powerful bidirectional diffusion priors to an efficient, unidirectional autoregressive paradigm. To overcome the attention distribution shift inherent in causal execution, we introduced a progressive three-stage distillation pipeline comprising Foundation Tuning, Teacher Forcing, and DMD. This architecture effectively bridges the gap between offline global processing and online continuous video editing, compressing the inference process to merely 4 steps. Furthermore, to alleviate spatial-temporal token redundancy, we proposed an AR-oriented Mask Cache mechanism. It reduces computational overhead while strictly guaranteeing zero visual degradation in unedited regions. Extensive quantitative and qualitative evaluations demonstrate that our framework achieves SOTA performance. It uniquely balances high-fidelity text alignment, robust structural preservation, and ultra-low latency, paving the way for highly practical, real-time streaming video editing applications.

%% file: sec/X_suppl.tex
\clearpage
\setcounter{page}{1}
\maketitlesupplementary

\tableofcontents
\setcounter{page}{1}
\renewcommand{\thepage}{S\arabic{page}}
\setcounter{section}{0}
\setcounter{figure}{0}
\setcounter{table}{0}

\appendix

\section{Discussion with Previous Methods}

Among existing works about video diffusion model~\cite{ma2025followfaster, ma2024followyouremoji, song2026vista, song2026streamingeffect, gao2026pai, song2024processpainter, Deng_2026_CVPR, deng2024compact, wu2025freeswim, wu2026vibe, liang2026spongebob, liang2026cot, xu2026smrabooth}, the two methods most similar to our proposed framework are Self-Forcing~\cite{huang2025self} and EgoEdit~\cite{li2025egoedit}. Here, We discuss the differences between these methods and our approach.

Self-Forcing is fundamentally designed for Text-to-Video generation scenarios, bridging exposure bias through self-generated conditioning. However, its reliance on an ODE initialization~\cite{yin2025slow, zhu2026causal} phase to establish an initial causal checkpoint introduces a severe computational bottleneck. While this ODE trajectory is manageable to sample from text in T2V tasks, constructing it for streaming video editing demands processing the original high-resolution and long sequence video data, incurring prohibitive computational overhead. Furthermore, T2V generation focuses on synthesizing from scratch, requiring only consistency with the previous generated content, while video editing necessitates strict spatial alignment with the input raw video. Directly applying the mechanism to editing tasks inevitably leads to structural deviation. In contrast, our proposed LiveEdit framework explicitly bypasses the costly ODE initialization by directly utilizing the autoregressive weights from Stage 2 as a highly stable distillation starting point. To strictly preserve source fidelity, we introduce an AR-oriented Mask Cache, prioritizing precise spatial restoration over free-form video synthesis.

EgoEdit pioneers streaming models for egocentric video editing, primarily validating the effectiveness of task-specific data within an established Self-Forcing pipeline. While it successfully demonstrates a continuous first-person visual dataset and benchmark, it structurally relies on this existing route without further inference acceleration for streaming video editing. Consequently, its research scope remains highly specialized and tailored to egocentric tasks, making it difficult to generalize to out-of-distribution tasks. Conversely, our LiveEdit is architected for general, high-fidelity streaming video editing across diverse scenarios. Instead of merely adapting data to existing pipelines, we introduce a comprehensive three-stage distillation pipeline that compresses inference to merely 4 steps. By dynamically decoupling computation, our framework achieves real-time responsiveness and strict background preservation in universal editing applications, breaking the limitations of single-domain adaptations.

\section{User Study}

\begin{figure*}[t]
    \centering
    \includegraphics[width=0.85\linewidth]{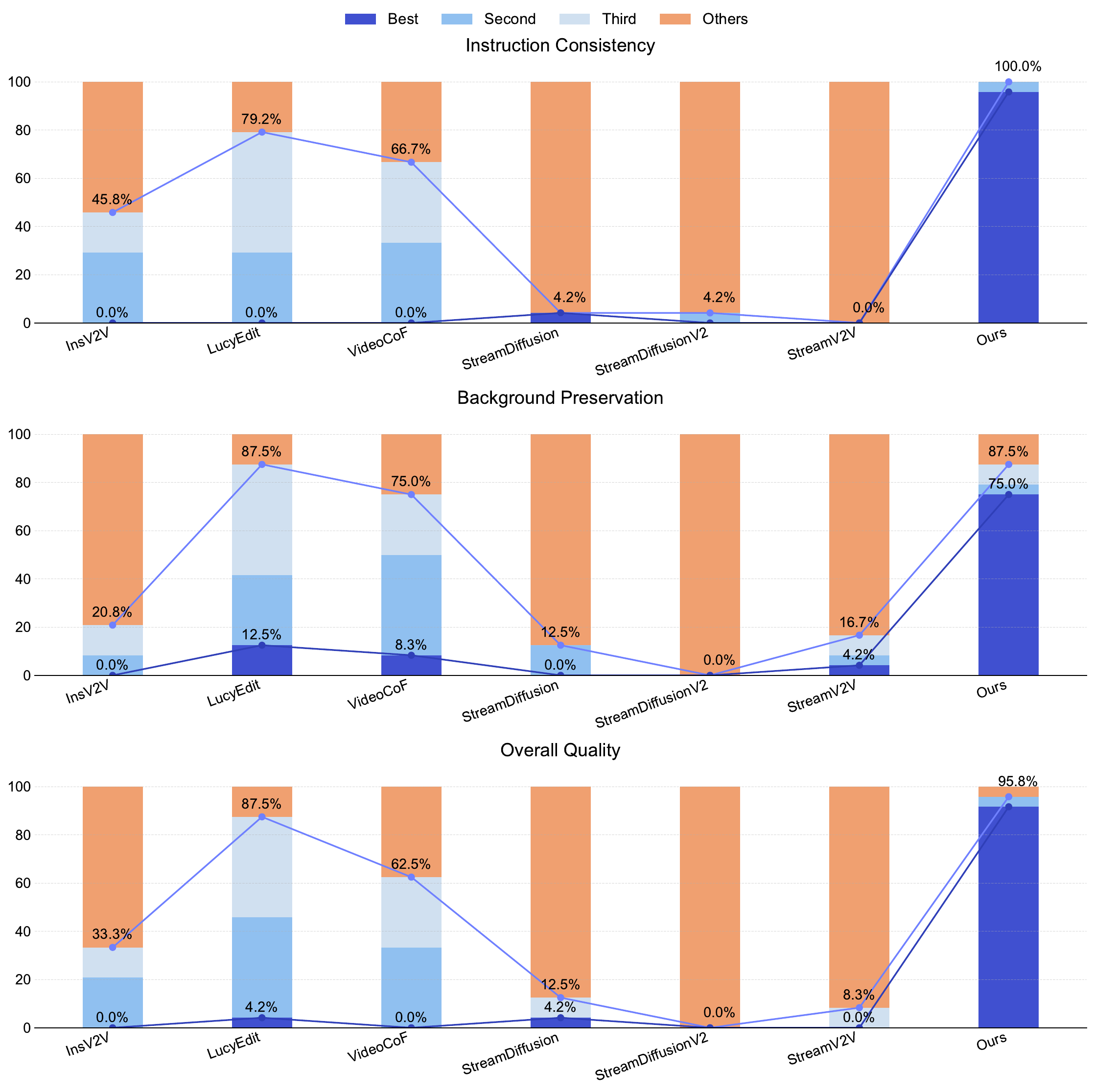}
\caption{\textbf{User study results.} Volunteers ranked the generated videos from our method and six baselines across three metrics: Instruction Consistency, Background Preservation, and Overall Quality. The line plots indicate the proportion of top-3 selections. Our proposed approach overwhelmingly dominates the evaluations, securing the vast majority of absolute "Best" rankings across all dimensions..}
    \label{fig:user_study}
\end{figure*}

To complement our quantitative metrics, we conducted a comprehensive user study to evaluate the perceptual quality of the generated videos. We invited 20 volunteers to review and rank the editing results from our method against six state-of-the-art baselines: InsV2V, LucyEdit, VideoCoF, StreamDiffusion, StreamDiffusionV2, and StreamV2V. Participants were asked to rigorously evaluate the videos based on three core dimensions: Instruction Consistency, Background Preservation, and Overall Quality. For each video group, volunteers ranked the methods, and we aggregated the results into "Best", "Second", "Third", and "Others" categories.

As illustrated in Fig.~\ref{fig:user_study}, our streaming editing framework demonstrates overwhelming superiority across all evaluation metrics. For Instruction Consistency, our method achieved a 100.0\% top-3 preference rate and monopolized the vast majority of the absolute "Best" rankings. In severe contrast, existing streaming generation baselines fall predominantly into the "Others" category, showing that they fundamentally struggled to execute the editing instructions accurately, .

Regarding Background Preservation, our approach received 75.0\% of the explicit "Best" votes and an 87.5\% top-3 preference rate. While offline bidirectional models like LucyEdit also exhibited competitive top-3 rates (87.5\%), they were rarely selected as the definitive top choice (only securing 12.5\% of "Best" votes) due to slight structural shifts or minor temporal inconsistencies. This stark contrast solidifies the effectiveness of our AR-oriented Mask Cache in strictly maintaining the visual integrity of unedited regions.

Ultimately, in terms of Overall Quality, our framework was consistently favored by the human evaluators, achieving a 95.8\% top-3 preference rate and definitively outperforming both offline and streaming baselines. The user study firmly aligns with our quantitative findings, confirming that our method uniquely balances high-fidelity semantic editing with strict spatial-temporal consistency.

\section{More cases}

\begin{figure*}[ht]
    \centering
    \includegraphics[width=1.\textwidth]{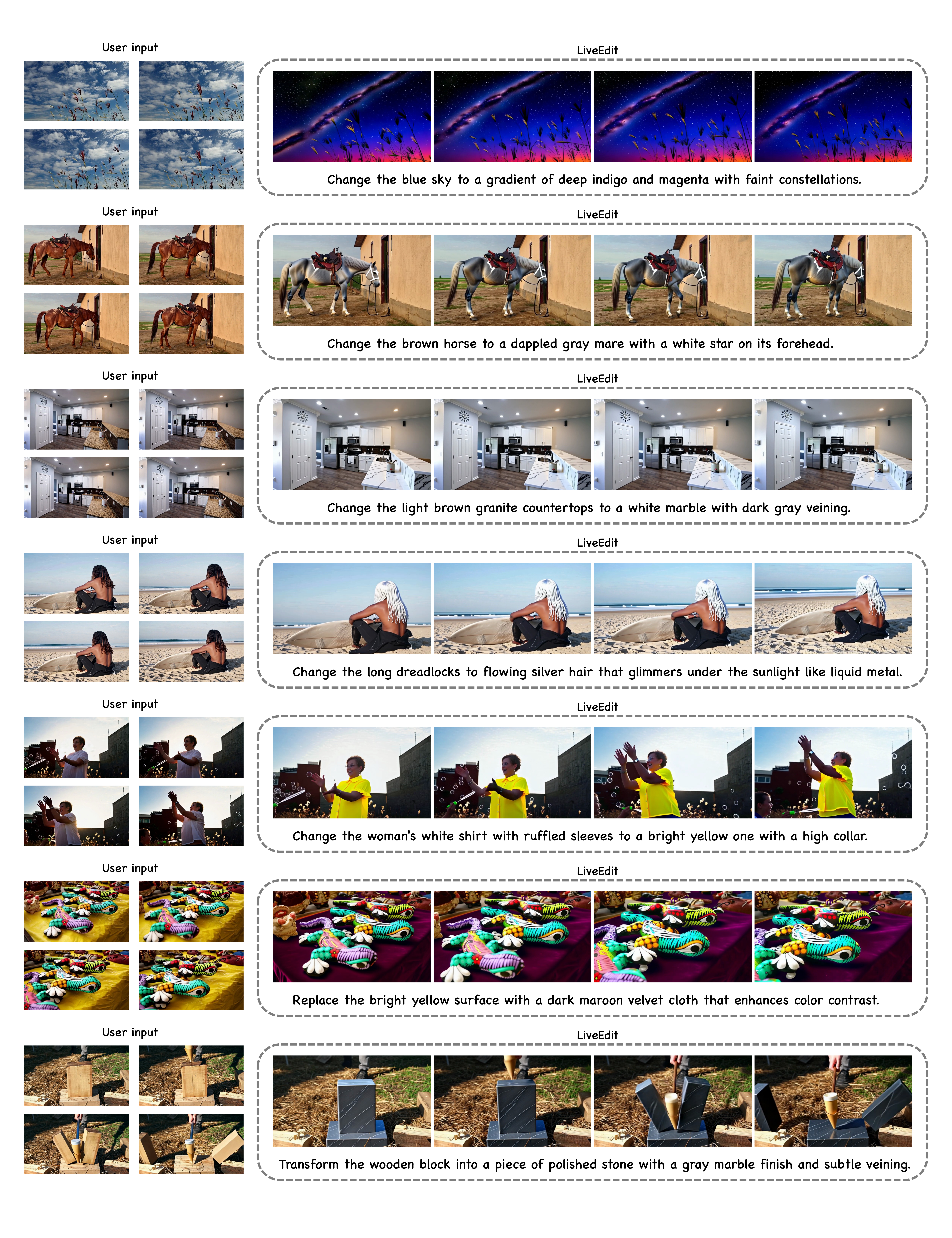}
    \caption{\textbf{More cases generated by LiveEdit}. 
    }
    \label{fig:case1}
\end{figure*}

\begin{figure*}[ht]
    \centering
    \includegraphics[width=1.\textwidth]{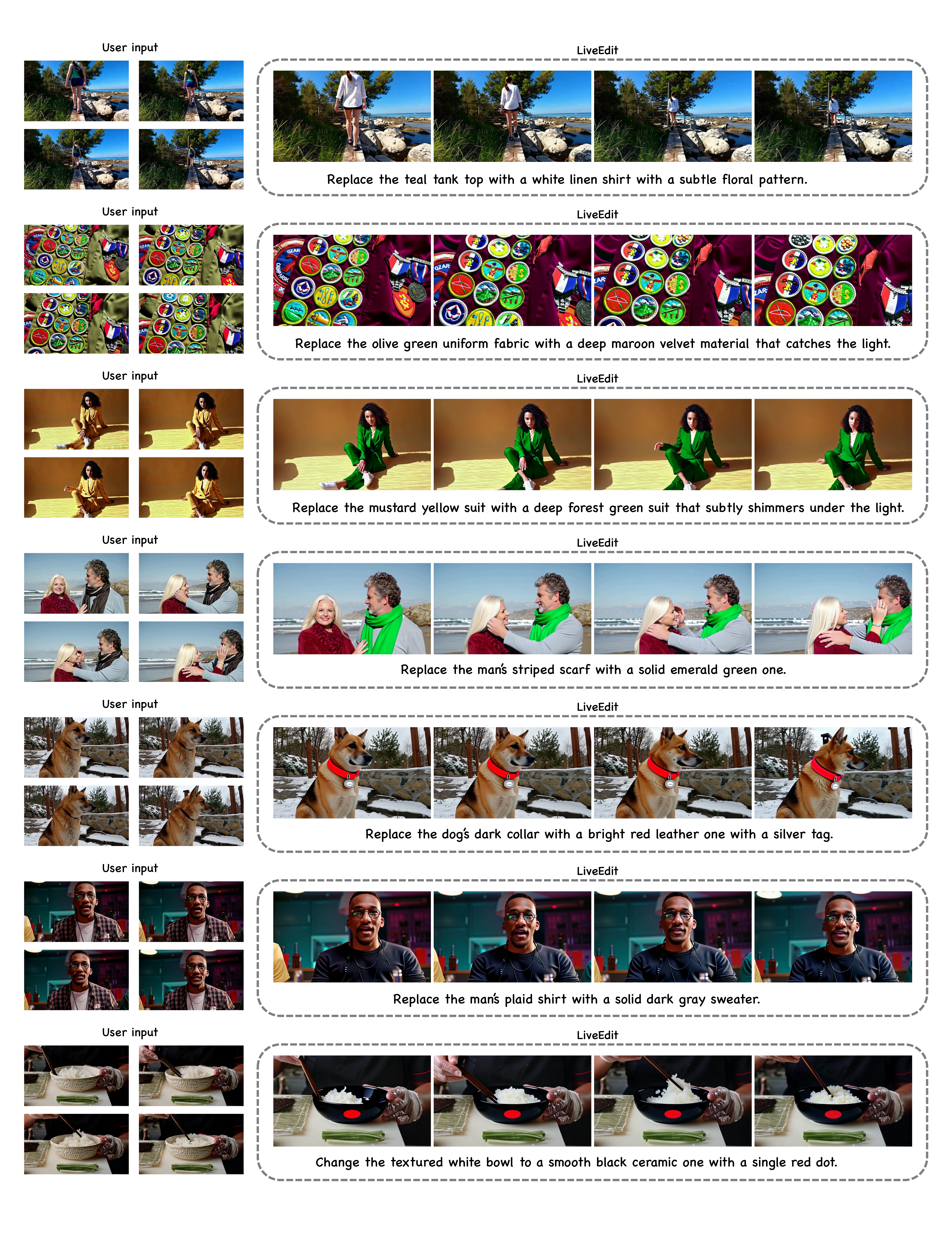}
    \caption{\textbf{More cases generated by LiveEdit}. 
    }
    \label{fig:case2}
\end{figure*}

\begin{figure*}[ht]
    \centering
    \includegraphics[width=1.\textwidth]{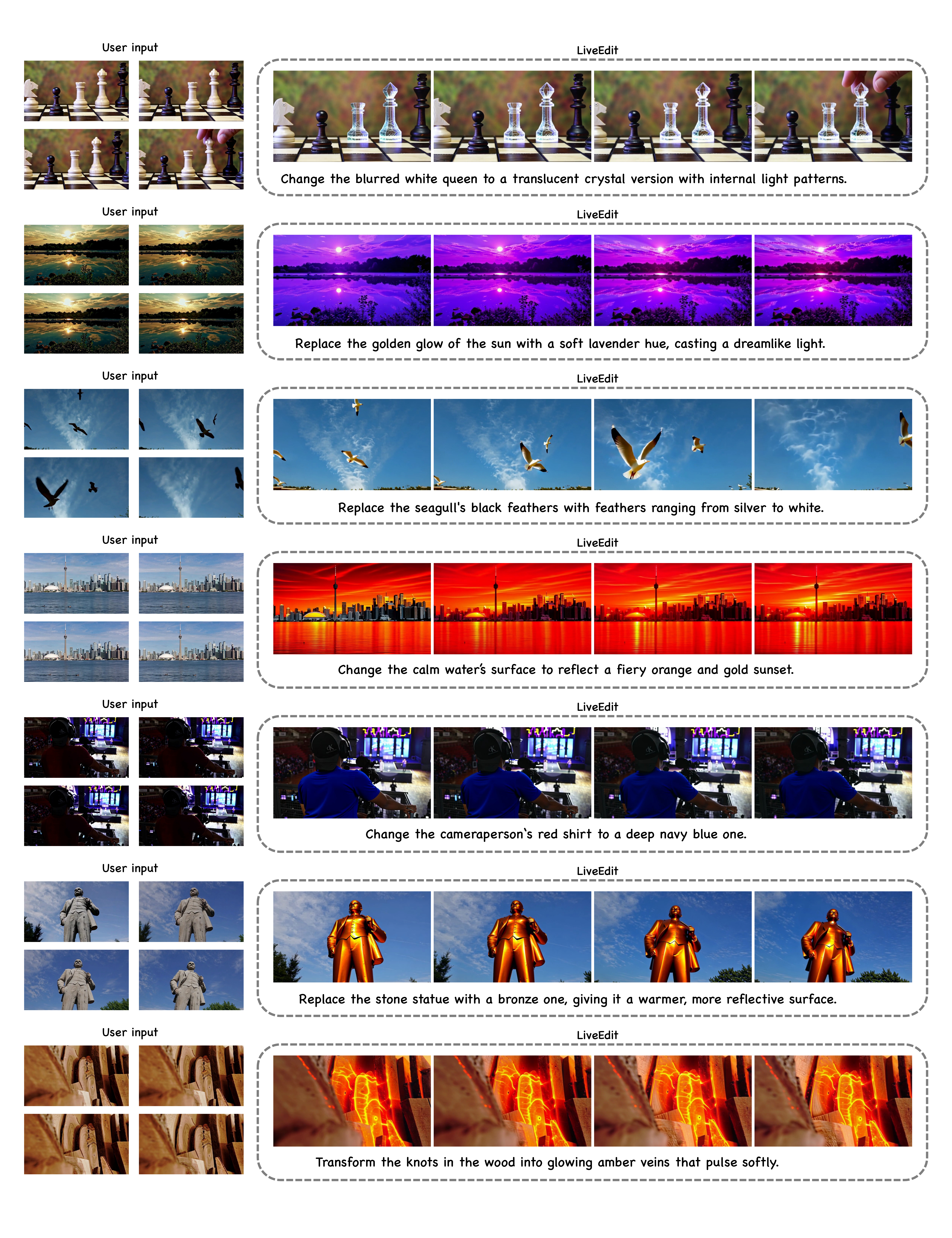}
    \caption{\textbf{More cases generated by LiveEdit}. 
    }
    \label{fig:case3}
\end{figure*}

We provide more cases to show the effectiveness of our method. (shown in Fig. \ref{fig:case1}, \ref{fig:case2}, \ref{fig:case3})

\section{More Comparisons}

\begin{figure*}[ht]
    \centering
    \includegraphics[width=1.\textwidth]{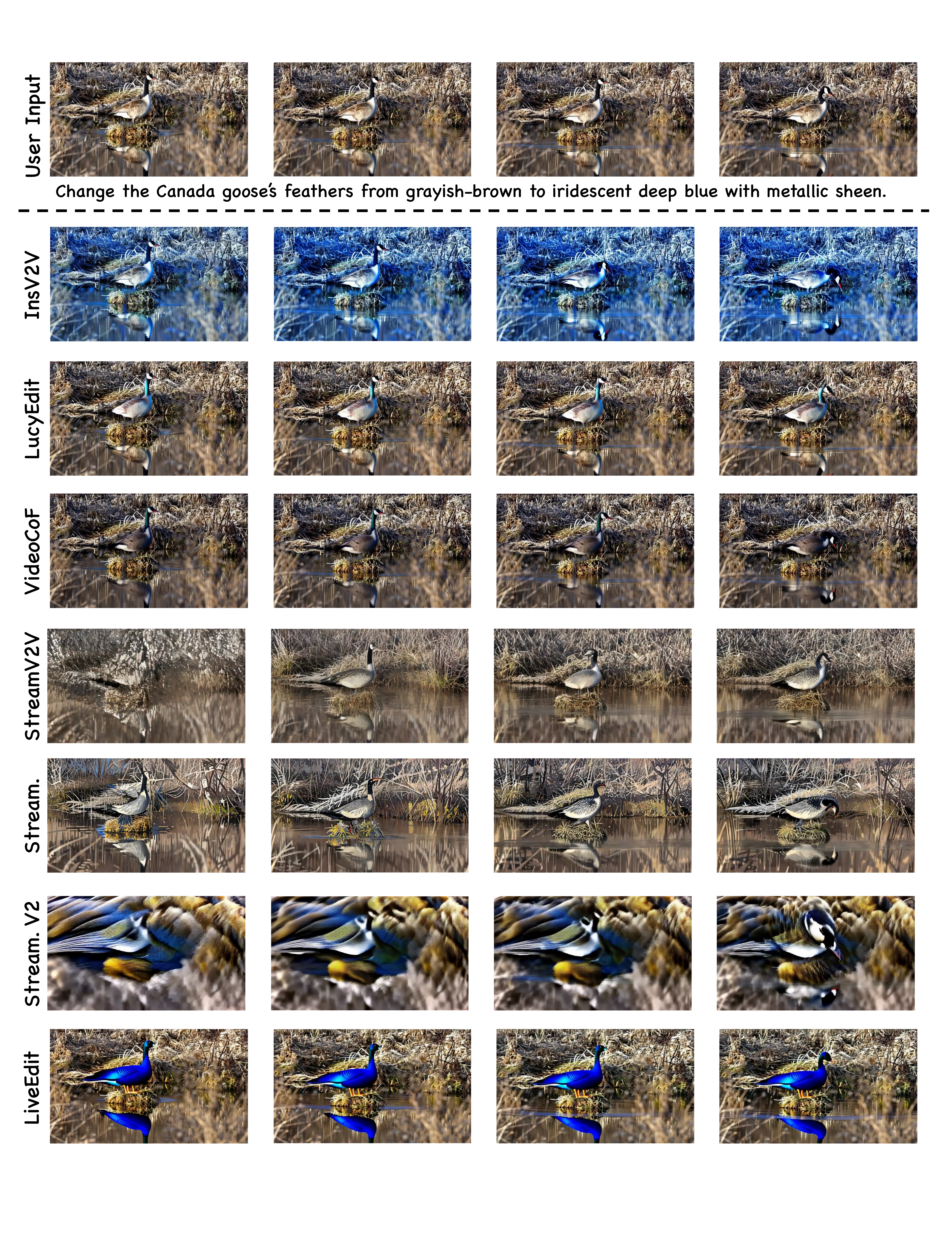}
    \caption{\textbf{More comparison between baseline and our LiveEdit}. 
    }
    \label{fig:case4}
\end{figure*}

\begin{figure*}[ht]
    \centering
    \includegraphics[width=1.\textwidth]{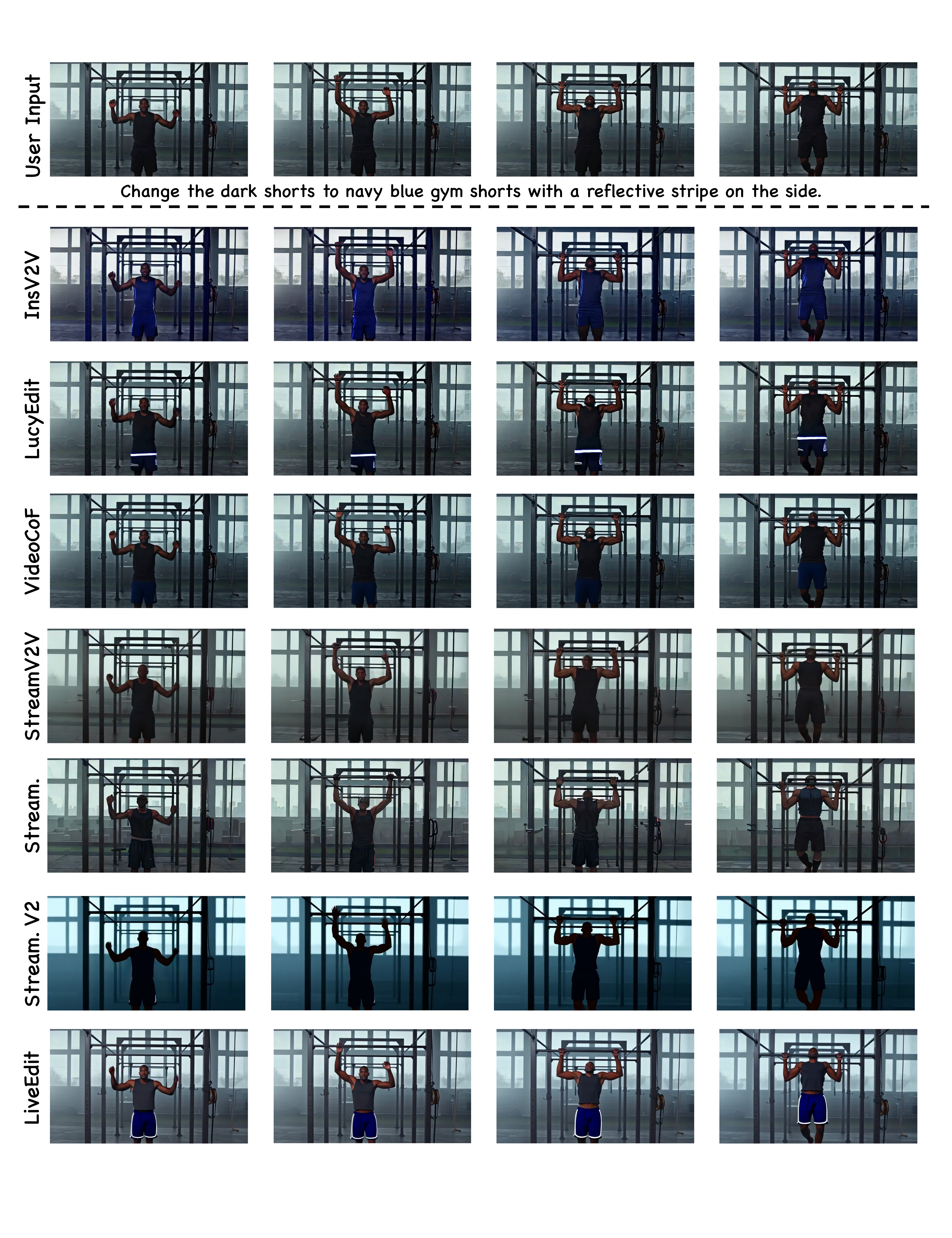}
    \caption{\textbf{More comparison between baseline and our LiveEdit}. 
    }
    \label{fig:case5}
\end{figure*}

We provide more comparisons to show the baseline and our method. (shown in Fig. \ref{fig:case4}, \ref{fig:case5})

